\documentclass{article}

\usepackage[preprint]{neurips_2022}

\usepackage[utf8]{inputenc} 
\usepackage[T1]{fontenc}    
\usepackage{hyperref}       
\usepackage{url}            
\usepackage{booktabs}       
\usepackage{amsfonts}       
\usepackage{nicefrac}       
\usepackage{microtype}      
\usepackage{xcolor}         
\usepackage{natbib}
\usepackage{xspace}
\usepackage{dsfont}
\usepackage{amsmath,amssymb}
\usepackage[ruled,norelsize]{algorithm2e}
\usepackage{multirow}
\usepackage{booktabs}
\usepackage{graphicx}
\usepackage{subcaption}
\usepackage{wrapfig}
\usepackage{enumerate}
\usepackage{enumitem}
\usepackage{url}
\usepackage{setspace}
\usepackage{arydshln}
\usepackage{hyperref}
\hypersetup{
    colorlinks=true,
    linkcolor=blue,
    filecolor=blue,      
    urlcolor=blue,
    citecolor=cyan,
}

\usepackage{amsmath,amsfonts,bm}

\newcommand{\eq}[1]{Eq.(\ref{#1})}
\newcommand{\anB}[1]{\langle #1 \rangle}

\def\vx{{\bm{x}}}

\DeclareMathAlphabet{\mathsfit}{\encodingdefault}{\sfdefault}{m}{sl}
\SetMathAlphabet{\mathsfit}{bold}{\encodingdefault}{\sfdefault}{bx}{n}

\def\cP{{\mathcal{P}}}
\def\cQ{{\mathcal{Q}}}

\def\cX{{\mathcal{X}}}

\newif\ifcomments
\commentstrue

\newcommand{\tblue}[1]{{\textcolor{blue}{#1}}}
\newcommand{\tred}[1]{{\textcolor{red}{#1}}}
\newcommand{\ttt}[1]{\texttt{#1}}

\newcommand{\method}{{\sc LogicLLaMA}\xspace}
\newcommand{\dataset}{{\sc Malls}\xspace}

\SetKwInput{KwInit}{Init}

\title{Harnessing the Power of Large Language Models for Natural Language to First-Order Logic Translation}

\author{Yuan Yang\textsuperscript{1}, Siheng Xiong\textsuperscript{1}, Ali Payani\textsuperscript{2}, Ehsan Shareghi\textsuperscript{3} \& Faramarz Fekri\textsuperscript{1}\\
\textsuperscript{1}Georgia Institute of Technology, 
\textsuperscript{2}Cisco,
\textsuperscript{3}Monash University\\
\ttt{\{yyang754@,sxiong45@,faramarz.fekri@ece.\}gatech.edu} \\
\ttt{apayani@cisco.com}\ \ \ 
\ttt{ehsan.shareghi}@monash.edu
}

\begin{document}

\maketitle

\begin{abstract}

Translating natural language sentences to first-order logic (NL-FOL translation) is a longstanding challenge in the NLP and formal logic literature.
This paper introduces \method, a LLaMA-7B model fine-tuned for NL-FOL translation using LoRA on a single GPU. \method is capable of directly translating natural language into FOL rules, which outperforms GPT-3.5. \method is also equipped to correct FOL rules predicted by GPT-3.5, and can achieve similar performance as GPT-4 with a fraction of the cost.
This correction ability was achieved by a novel supervised fine-tuning (SFT) + reinforcement learning with human feedback (RLHF) framework, which initially trains on synthetically perturbed NL-FOL pairs to encourage chain-of-thought reasoning and then fine-tunes with RLHF on GPT-3.5 outputs using a FOL verifier as the reward model.

To train \method, we present \dataset (large language \textbf{M}odel gener\textbf{A}ted N\textbf{L}-FO\textbf{L} pair\textbf{S}), a dataset of 34K high-quality and diverse sentence-level NL-FOL pairs collected from GPT-4. The dataset was created by implementing a pipeline that prompts GPT-4 for pairs, and dynamically adjusts the prompts to ensure the collection of pairs with rich and diverse contexts at different levels of complexity, and verifies the validity of the generated FOL rules. Codes, weights, and data are available at \href{https://github.com/gblackout/LogicLLaMA}{https://github.com/gblackout/LogicLLaMA}.

\end{abstract}

\section{Introduction}
Large language models (LLMs) have established state-of-the-art results on several reasoning and generation benchmark tasks~\citep{DBLP:journals/corr/abs-2303-08774,DBLP:journals/corr/abs-2204-02311}. Despite their success, LLMs struggle with logical reasoning (a prime example of System~2 task \citep{kahneman2011thinking}), or maintaining logical consistency during generation~\citep{DBLP:conf/nips/NyeTTL21}. The common denominator of both is the absence of explicit logical grounding which could impose the consistency of a generated output and the state of the world (i.e., premises of the reasoning task, or the previously generated text). While desired, the existing tools and systems that foster such explicit grounding~\citep{abzianidze-2017-langpro, bos-markert-2005-recognising} of Natural Language (NL) are brittle, and rely on hard-coded First-Order Logic~(FOL) rules and facts, which is impractical for real-world use. 

Recent variants of LLMs (i.e., GPT-4) exhibit impressive few-shot capabilities in NL-FOL translation tasks. This rapid improvement comes after recent observations~\citep{han_folio_2022} which highlighted major defects of previous LLMs (e.g., GPT-3 davinci).  Nonetheless, even the most powerful LLMs to this date cannot solve the NL-FOL translation task entirely, and for complex NL statements, they typically generate an answer which still requires a few ``corrections''.  However, in the absence of fine-tuning option (not available for RLHF-trained LLMs), most of the heavy lifting in this translation task is offloaded on the few-shot examples and prompt engineering. Not to mention, the cost element of using an LLM as a dedicated tool (or fine-tuning them) for NL-FOL translation could be prohibitive. 

In order to improve the translation quality of LLMs (i.e., GPT-3.5), we present a framework that runs every output from GPT-3.5 through a small language model~(\method), a LLaMA-7B model~\citep{touvron2023llama} for NL-FOL translation fine-tuned with LoRA~\citep{hu2021lora}. \method is trained to correct outputs from GPT-3.5 (through an iterative correction) while also being able to act as a standalone direct NL-to-FOL translator. For training \method, we collected a high-quality and diversified dataset of 34K sentence-level NL-FOL pairs from GPT-4. We then created a perturbed version of the FOL in each pair to produce a controlled perturbation dataset, where each perturbed pair is accompanied by a ``correction instruction'' to undo the perturbation. We propose a novel SFT+RLHF framework that first trains \method on the synthetically perturbed NL-FOL pairs, equipping \method with generating corrective prompts, and then fine-tunes it with RLHF on the GPT-3.5 outputs using a FOL verifier as the reward model.

In our experiments, we probe the capabilities of the most recent LLMs in both zero- and few-shot settings in the NL-FOL translation task on two  benchmarks with different levels of complexity, LogicNLI~\citep{tian-etal-2021-diagnosing} and FOLIO~\citep{han_folio_2022}. We highlight, on the challenging dataset of FOLIO, the latest GPT-3.5 with 5-shot examples in the prompt does not go above 0.767 logical equivalence (LE) score, our proposed approach could iteratively improve its performance and reduce the gap between GPT-3.5 and GPT-4 (i.e., GPT-3.5+\method achieves 0.849 LE compared with GPT-4 score of 0.855)
with a fraction of the cost.\footnote{As of May 2023, GPT-3.5 costs \$0.002/1K tokens whereas GPT4 costs \$0.03/1K for prompt and \$0.06/1K for completion.} Additionally, we demonstrate \method capabilities as a standalone model for NL-FOL translation task, outperforming GPT-3.5 on both FOLIO and LogicNLI, while being highly competitive with GPT-4.





\section{Related Work}


\textbf{NL-FOL translation}.
Natural language to first-order logic (NL-FOL) translation is a critical task that serves as the foundation of a wide range of logic-backed NLP applications, such as textual entailment~\citep{bos-markert-2005-recognising}, NL inference~\citep{angeli2014naturalli} and theorem proving~\citep{polu2020generative}.
Traditionally, NL-FOL translation has been addressed via rule-based methods~\citep{abzianidze-2017-langpro,DBLP:conf/uai/ZettlemoyerC05,bos-markert-2005-recognising}. Due to the complexity of natural language, these methods are difficult to scale to real-world applications.
Recently, there has been an increasing interest in approaching this task via neural approaches~\citep{lu-etal-2022-parsing, cao-etal-2019-semantic,hahn2022formal,wang2021logic,singh2020exploring,levkovskyi2021generating}.
The recent release of powerful LLMs such as GPT-3.5 and GPT-4 gives rise to a new paradigm: using LLMs to perform the bulk of the translation task, thereby benefiting from their generalization capabilities and capacity to handle complex and diverse language constructs.
In this work, we investigate this paradigm and propose to collect NL-FOL pairs from GPT-4 and fine-tune a LLaMA-7B model on it.

\textbf{NL-FOL datasets}. 
Many datasets that focus on logical reasoning ability have been proposed recently. For example, LogiQA~\citep{liu2020logiqa}, RuleTaker~\citep{clark2020transformers}, ReClor~\citep{yu2020reclor} and text2log~\citep{levkovskyi2021generating}.
However, these datasets either do not provide sentence-level FOL annotations, or the annotations are generated without verification.
Among these works, LogicNLI~\citep{tian-etal-2021-diagnosing} and FOLIO~\citep{han_folio_2022} are closest to our work, which provides NL statements with parallel FOL annotations. 
However, pairs in LogicNLI are generated synthetic and share a similar FOL template.
FOLIO consists of real-world expert-written pairs, but the size of 2K is insufficient for fine-tuning an LLM.
This work extends the prior work and proposes to collect ``silver'' NL-FOL pairs from GPT-4. As a result, \dataset has collected 34K pairs that are more diverse in terms of context and complexity. In experiments, we use LogicNLI and FOLIO as the ``gold'' sets to evaluate the LLM fine-tuned on \dataset
and demonstrate that it is of high quality.



\section{\dataset Dataset Creation}

We create the \dataset dataset by collecting NL-FOL pairs from GPT-4 which is considered to be the most powerful LLM to date. As of May 2023, \dataset has reached the size of 34K and we plan to continue expanding the dataset in future versions.

\textbf{Motivation}. 
One of the goals to create such a dataset is to provide a corpus for fine-tuning and evaluating NL-FOL translation models. However, one may ask ``\textit{if \dataset is to be generated by yet another LLM, i.e., GPT-4, then why shouldn't one use GPT-4 for the task already?}'' The motivation lies in the cost and privacy. While GPT-4 yields state-of-the-art performance, its API access is costly and not entirely publicly available to date; on the other hand, institutes and companies may have sensitive data that cannot be shared with a third party and they want to deploy a local LLM with similar performance. In \S\ref{sec:logicllama}, we show this can be achieved by fine-tuning a LLaMA-7B model on \dataset on a single GPU. 

\begin{table}[t]
\centering
\caption{Statistics of \dataset, LogicNLI, and FOLIO datasets.}
\label{tab:dataset_stats}
\resizebox{\textwidth}{!}{%
\begin{tabular}{@{}lcccccccccccccc@{}}
\toprule
\multicolumn{1}{c}{\multirow{2}{*}{Dataset}} & \multirow{2}{*}{Source}                                   & \multirow{2}{*}{\begin{tabular}[c]{@{}c@{}}\#NL-FOL\\ pairs\end{tabular}} & \multicolumn{2}{c}{NL}                                                                                        &  & \multicolumn{9}{c}{FOL}                                                                                                                              \\ \cmidrule(l){4-15} 
\multicolumn{1}{c}{}                         &                                                           &                                                                           & \begin{tabular}[c]{@{}c@{}}Vocab\\ size\end{tabular} & \begin{tabular}[c]{@{}c@{}}Avg.\\ \#words\end{tabular} &  & \begin{tabular}[c]{@{}c@{}}Avg.\\ \#literals\end{tabular} & $\forall$ & $\exists$ & $\neg$ & $\land$ & $\lor$ & $\to$ & $\leftrightarrow$ & $\oplus$ \\ \midrule
FOLIO$^3$                                    & Expert & 2K                                                                        & 5105                                                 & 10.4                                                   &  & 2.1                                                       & 1111      & 182       & 421    & 631     & 167    & 1137  & 17                & 121      \\
LogicNLI$^3$                                 & Synthetic                                                 & 12K                                                                       & 2061                                                 & 13.9                                                   &  & 2.8                                                       & 2783      & 5327      & 10230  & 6590    & 2373   & 8712  & 3288              & 0        \\
\dataset                                     & GPT-4                                                     & 34K                                                                       & 22715                                                & 16.1                                                   &  & 4.6                                                       & 32865     & 2036      & 4567   & 30143   & 6402   & 30667 & 3726              & 2150     \\ \bottomrule
\end{tabular}%
}
\end{table}
\begin{figure}[t]
    \centering
    \includegraphics[angle=-90, trim={0.5cm 0 26cm 0cm},clip,scale=0.5]{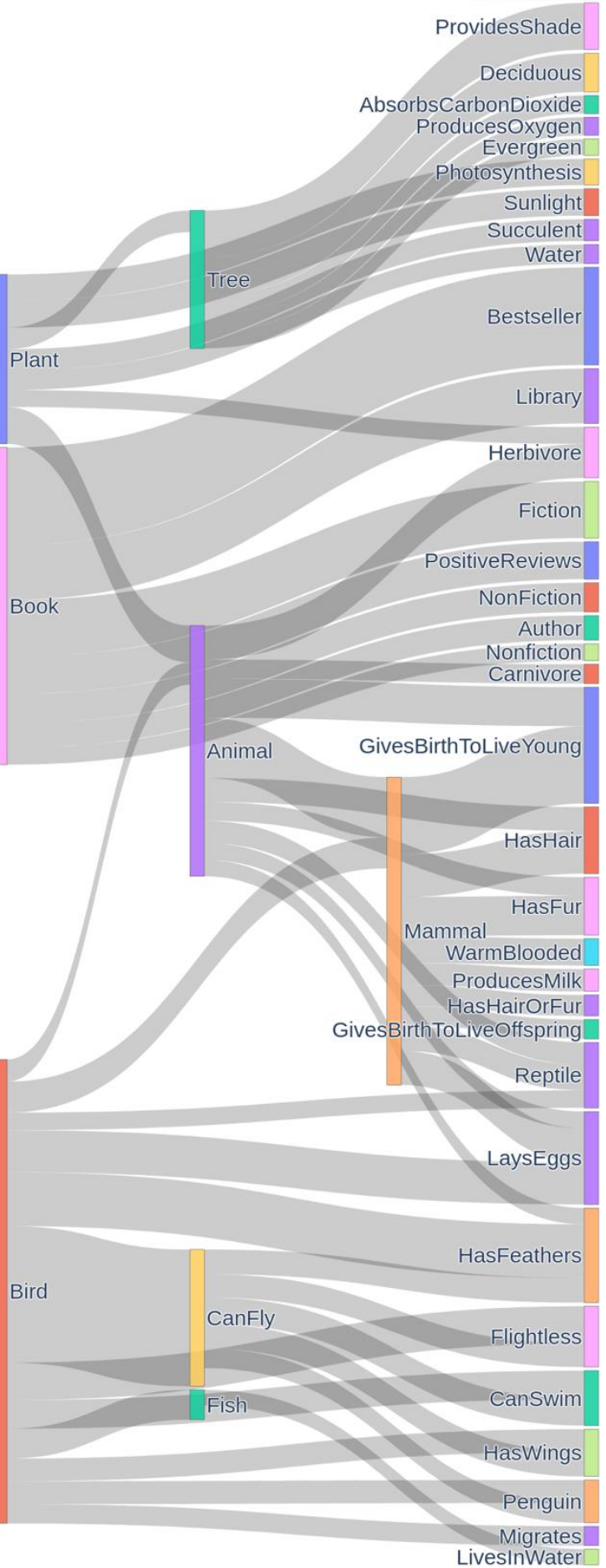}
    \caption{Snippet from the top 200 frequent FOL term pairs in \dataset (for full version see Appendix~\ref{app:dataset_app}). Many terms are associated with a wide range of other terms, which suggests the rules are semantically and contextually diverse.}
    \label{fig:sankey-small}
\end{figure}

\subsection{Prompt pipeline}

To collect data from GPT-4, we implemented a prompting pipeline that dynamically adjusts the prompts to both ensure the \emph{diversity} and \emph{validity} of the NL-FOL pairs. The pipeline consists of the following modules: (1) \textbf{N-gram frequency counter}; (2) \textbf{Prompter}; and (3) \textbf{FOL rule verifier}.

\textbf{N-gram frequency counter}. 
During prompting, we keep track of the frequencies of the N-grams in the entire NL statement corpus. Specifically, we track 1- and 3-grams. Once the frequency of a specific N-gram in the collected data reaches the frequency threshold (500 and 250 respectively),
 we will instruct GPT-4 to not produce any NL-FOL pairs including it. For example, 
``\textit{... DO NOT involve concepts and terms (and the synonyms) such as animal, food, ...''}. The list of N-grams in the instruction grows as more reach the frequency threshold.

\textbf{Prompter}. 
A prompter assembles the prompts generated from different modules (prompt table shown in Appendix~\ref{app:dataset_app}): 
(1) \textsc{system prompt}: specifying the basic requirements such as the syntax and generation format.
(2) \textsc{few-shot examples prompt}: consisting 5 NL-FOL pair examples randomly sampled from the corpus. Initially, pairs are sampled from the FOLIO dataset
and later on from the GPT-4-generated ones (we checked to ensure none of the FOLIO examples, or close variations are leaked into the GPT-4 generated NL-FOL pairs.). 
This diversifies the prompts and leads to less similar examples 
.(3) \textsc{negative N-gram prompt}: instructing GPT-4 not to involve frequent N-grams (introduced earlier) in the generated NL-FOL pairs. (4)\textsc{FOL prompts}: generating prompts that specify the desired form of FOL rules, i.e., the number of variables and whether or not to include more logic operators such as $\oplus$, $\neg$, and $\lor$ which we found GPT-4 tends to ignore in default generation. These configurations are picked randomly every time the prompt is generated. (5) \textsc{break-down prompt}: We found GPT-4 by default tends to make over-complicated predicates that absorb important logical meanings. For example,
``
\#\#\# NL:
\textit{A fruit is considered ripe if it is mature and its color has changed from green to red.}
\#\#\# FOL:
$\forall x (\ttt{Fruit}(x) \land \ttt{Mature}(x) \land \ttt{ColorChangedToRed}(x) \to \ttt{Ripe}(x))$.
''
The predicate \ttt{ColorChangedToRed} is complicated and should be broken down into ``$\ttt{ColorBefore}(x, y) \land \ttt{ColorAfter}(x, z) \land \ttt{Green}(y) \land \ttt{Red}(z)$''.
We alleviate this by detecting long predicate names and including a prompt encouraging the model to break down the rules.

\textbf{FOL rule verifier}.
GPT-4 can sometimes generate syntactically invalid FOL rules. We implement a verifier that checks the syntax of the rules. Specifically, we specify the context-free grammar (CFG) of the expected FOL rule 
and parse the generated FOL with NLTK~\footnote{https://www.nltk.org/} CFG parser, and erase those that could not be parsed (grammar and example parse trees in Appendix~\ref{app:dataset_app}).

\subsection{Dataset statistics}

\textbf{General statistics}.
We show the general statistics in Table~\ref{tab:dataset_stats} together with those of LogicNLI and FOLIO\footnote{Note that the FOLIO statistics are different from those reported in~\citep{han_folio_2022}. As of May 2023, the released dataset misses the ground truth FOL annotations for conclusions in the training set, and some pairs contain duplicates and invalid FOL rules. We removed those during pre-processing.
Also, the LogicNLI statistics are obtained from the official repo \href{https://github.com/omnilabNLP/LogicNLI}{here}, which contains 12K samples instead of the 20K reported in the paper.
}.
\dataset contains 34K NL-FOL pairs, which is significantly larger than LogicNLI and FOLIO, and different from LogicNLI which is synthetically generated, the pairs are also more diverse and contextually rich, where the NL statements have a vocabulary size of 22.7K and an average length of 16 compared to 10 in FOLIO. For FOL rules, the average number of literals reached 4.6 indicating more complex rules (also see Figure~\ref{fig:literal-freq} in Appendix~\ref{app:dataset_app}).

\textbf{Pair diversity}.
The NL-FOL rules in \dataset are highly diverse. To see this, we investigate the frequencies and the correlations of the FOL \textit{terms}. A \emph{term} is either a predicate name or a named entity in a FOL rule. For example, ``$\forall x ((\ttt{Person}(x) \land \ttt{Drinks}(x)) \to \ttt{DependentOn}(x, \ttt{Caffeine}))$'' consists of 4 terms, i.e., \ttt{Person}, \ttt{Drinks}, \ttt{DependentOn} and \ttt{Caffeine}. \dataset has a total term vocabulary size of 49394 and the most frequent terms occur less than 2K times (Figure~\ref{fig:term-freq} in Appendix~\ref{app:dataset_app}), suggesting a diverse vocabulary distribution. On the other hand, we investigate the correlations between terms and illustrate the top 200 frequent term pairs. We show a snippet of this in Figure~\ref{fig:sankey-small} (for the full version, see Figure~\ref{fig:sankey} in Appendix~\ref{app:dataset_app}). Note that if a term is associated with many other terms, this typically means the rules involving that term are diverse in semantics and context, and Figure~\ref{fig:sankey-small} suggests that it is indeed the case. For example, for rules involving \ttt{Book}, they cover the knowledge of its genre (e.g., \ttt{Fiction}), places (e.g., \ttt{Library}), viewership (e.g., \ttt{Bestseller} and \ttt{PositiveReviews}), and so on.

\textbf{NL-FOL alignment}.
Apart from checking the FOL validity, we also implemented a simplistic verifier that checks the alignment between the NL statement and the FOL rule. This is done by treating the FOL as a query and computing its term frequency in the NL, and then rejecting those that are below a threshold. 
Apart from this, we did not conduct a rigorous alignment check in the creation of \dataset. 
In fact, the best way to date to ensure alignment correctness is checking them manually as that in FOLIO dataset creation. 
This is prohibitive to do for a dataset of this size for an academic budget.
That said, we recommend treating the dataset as ``silver'' labels and using it for training, and using another dataset with ``gold'' labels for evaluation.
Nevertheless, in \S\ref{sec:experiments}, we demonstrate that a model trained solely on this silver dataset can still achieve a similar performance as GPT-4, when evaluated on the gold sets such as FOLIO and LogicNLI.





\section{Fine-tuning \method for NL-FOL Translation}
\label{sec:logicllama}

In this section, we discuss how to fine-tune the LLaMA-7B~\citep{touvron2023llama} model on the \dataset to reach a GPT-4 level performance. We refer to this model as \method. Throughout the remainder of this section, we will refer to the silver FOLs in \dataset as ground truth.

Unlike typical NLP tasks, where one fine-tunes it with a task-agnostic objective such as autoregression, fine-tuning for NL-FOL translation is nontrivial. Specifically, we address the following challenges: 

\textbf{(C1) What is the input and output of the \method? And how to prepare the training data from \dataset?}
In \S\ref{sec:naive-method}, we first consider the naive approach, where \method is trained to predict the correct FOL directly. While it does not need any additional data other than the original \dataset, it yields sub-optimal performance.
We found better performance is achieved by eliciting the chain-of-thought (CoT) steps and gradually correcting the FOL predicted by another model, e.g., GPT-3.5. 
But, such training requires the ground-truth CoT steps which are not available in \dataset. In \S\ref{sec:cot-rlhf}, we propose to address this by first fine-tuning the model on synthetically perturbed FOLs with ground-truth CoT steps and then conducting the RLHF to correct the real outputs of GPT-3.5. 

\textbf{(C2) How to evaluate the generated FOL rules?} 
Consider two FOL rules (denoted as $R$ and $R'$)  generated from an LLM ``$R: \neg(P(A) \land P(B))$'' and ``$R': \neg P(A) \lor \neg P(B)$'' --- $R$ and $R'$ are logically equivalent but are different in the text; also consider a pair of rules ``$R:\forall x P(x)$'' and ``$R': \forall x \forall y P(x) \land Q(y)$''--- if $R$ is the ground-truth and $R'$ the LLM prediction, how should one measure the distance and supervise the model? We address this in \S\ref{sec:fol-eval-reward-le}.

\begin{figure}[t]
    \centering
    \includegraphics[trim={0 3.4cm 0cm 0},clip,width=0.8\linewidth]{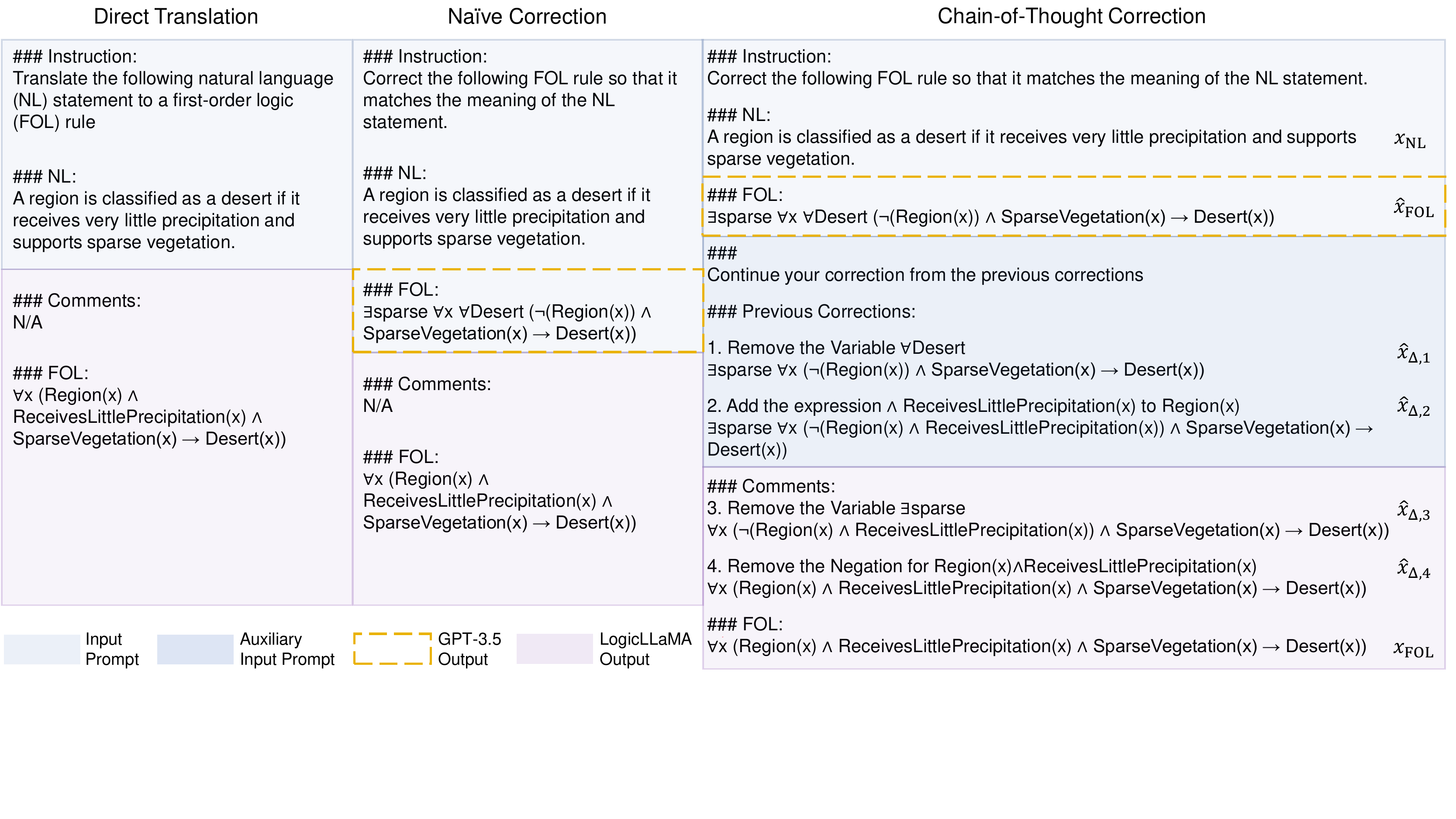}
    \caption{Input and expected outputs for direct translation, naive correction, and CoT correction.}
    \label{fig:three_modes}
\end{figure}

\subsection{Fine-tuning for direct translation and naive correction}
\label{sec:naive-method}

The \method can be trained to directly translate the FOL from NL, which we refer to as \textbf{(T1) direct translation} task; it can also be trained to \textit{correct} the generated FOL from a more powerful model such as GPT-3.5, which we refer to as the correction task. 
In this section, we consider the \textbf{(T2) naive correction} approach, where the correction is done in one go.
The intuition is that we found in experiments GPT-3.5 is good at doing the ``heavy-lifting'' part of the translation and can capture the main part of the FOL rule; then presumably, one can train a smaller model that corrects the output from the GPT-3.5 to get a better result. 

We train both \textbf{(T1)} and \textbf{(T2)} via standard autoregression objective. Specifically, we fine-tune a LLaMA-7B model with LoRA (for all the attention weight matrices) on \dataset.  
The left two columns in Figure~\ref{fig:three_modes} show the input and output sequence of the two tasks: 
let $\anB{\vx_\text{NL}, \vx_\text{FOL}}$ be an NL-FOL pair from \dataset;
for \textbf{(T1)}, the input and output are the original sequences $\vx_\text{NL}$ and $\vx_\text{FOL}$ respectively;
and for \textbf{(T2)}, let $\hat{\vx}_\text{FOL} = \text{GPT}(\vx_\text{NL})$ be the FOL predicted by GPT-3.5, the input is the NL and the prediction 
put together $[\vx_\text{NL}, \hat{\vx}_\text{FOL}]$ and the output is the ground-truth FOL, $\vx_\text{FOL}$. 

\subsection{Chain-of-Thought correction via SFT and RLHF}
\label{sec:cot-rlhf}

\begin{figure}[t]
    \centering
    \includegraphics[trim={0 4cm 4.8cm 0},clip,width=0.8\linewidth]{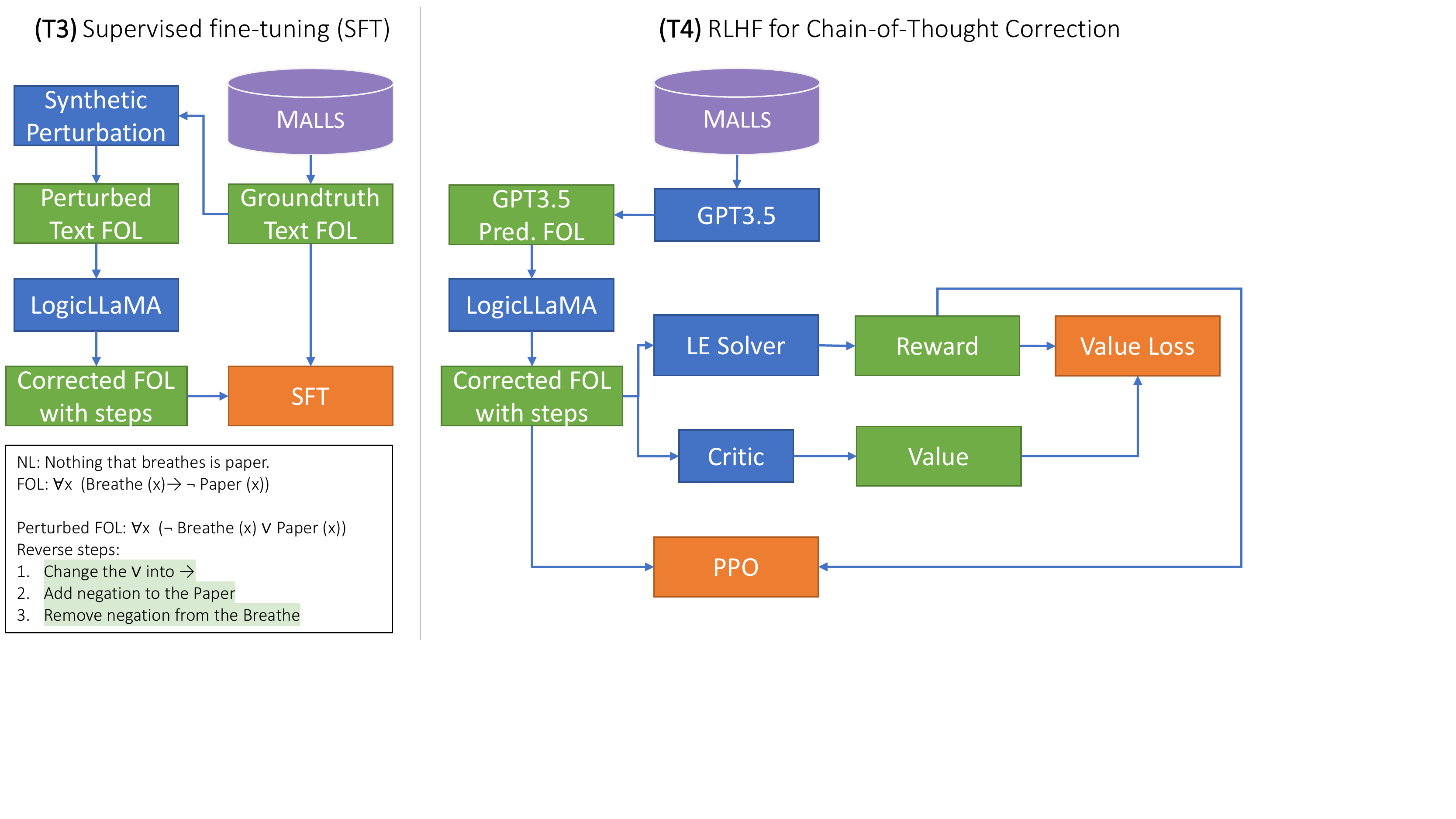}
    \caption{Overview of the SFT and RLHF training for the Chain-of-Thought (CoT) correction mode of \method.}
    \label{fig:sft-rlhf-overview}
\end{figure}

While \textbf{(T1)} direct translation and \textbf{(T2)} naive correction are easy to train, they do not lead to optimal performance. 
Inspired by the Chain-of-Thought (CoT) technique~\citep{wei2022chain}, we found that training the model to produce the intermediate steps during the correction often leads to better performance.
Such examples are shown in Figure~\ref{fig:correction-examples}.

To train such a model, one needs a dataset consisting of not only the ground-truth $\anB{\vx_\text{NL}, \vx_\text{FOL}}$, but also the CoT steps specific to a predicted FOL. Formally, recall that $\hat{\vx}_\text{FOL}$ is the predicted FOL by GPT-3.5, then we need the ground-truth steps $\hat{\cX}_\Delta = [\hat{\vx}_{\Delta,1}, \hat{\vx}_{\Delta,2}, ..., \hat{\vx}_{\Delta,T}]$, such that they form a valid CoT sequence $[\hat{\vx}_\text{FOL}, \hat{\vx}_{\Delta,1}, \hat{\vx}_{\Delta,2}, ..., \hat{\vx}_{\Delta,T}, \vx_\text{FOL}]$.
For example, the right column of Figure~\ref{fig:three_modes} shows a 4-step CoT correction. 

However, as stated in \textbf{(C1)}, we do not have ground-truth CoT steps $\hat{\cX}_\Delta$ for the predicted FOL from GPT-3.5. We propose to address this issue using a combination of supervised fine-tuning (SFT) on a \emph{synthetically perturbed}   dataset with ground-truth CoT steps, and reinforcement learning with human feedback (RLHF) training on the real GPT-3.5 output with a logical equivalence solver (discussed in \S\ref{sec:fol-eval-reward-le}) as the reward model.

Specifically, we refer to the SFT step as \textbf{(T3) SFT CoT Correction}. And as shown in the left column of Figure~\ref{fig:sft-rlhf-overview}, we create a synthetic FOL dataset by perturbing the ground-truth FOL rule and obtaining the ground-truth CoT steps by reversing the past perturbations.
And, we refer to the RLHF step as \textbf{(T4) RLHF CoT Correction}, which is shown in the right column of Figure~\ref{fig:sft-rlhf-overview}.

\subsubsection{FOL Rule Perturbations and SFT}

\begin{figure*}[t]
    \begin{minipage}{\textwidth}
    \begin{minipage}{0.59\textwidth}
        \vspace{0pt}
        \centering
        \captionof{table}{The list of all atomic perturbations.}
        \label{tab:perturb-operations}
        \resizebox{\columnwidth}{!}{%
        \begin{tabular}{@{}llll@{}}
        \toprule
        \multicolumn{1}{l}{Operation Type} & \multicolumn{1}{c}{Subtypes} & \multicolumn{1}{c}{Original}                             & \multicolumn{1}{c}{Perturbed}                     \\ \midrule
        \multirow{4}{*}{Label Change}      & Change Predicate             & $\tblue{P}(A) \land R(B)$                                & $\tred{R}(A) \land R(B)$                          \\\cdashline{3-4}
                                           & \multirow{2}{*}{Change Term} & \multirow{2}{*}{$\forall \tblue{x}\;\; P(x) \land P(B)$} & $\forall \tred{y} \;\; P(x) \land P(B)$           \\
                                           &                              &                                                          & $\forall x \;\; P(x) \land P(\tred{x})$     \\\cdashline{3-4}
                                           & Change Operator              & $\forall x \;\; P(x) \;\tblue{\land}\; P(B)$                 & $\forall x \;\; P(x) \;\tred{\lor}\; P(B)$            \\ \midrule
        \multirow{4}{*}{Insert}            & \multirow{2}{*}{Insert Term} & \multirow{2}{*}{$\forall x \;\; P(x) \land P(B)$}        & $\forall x \; \tred{\exists y} \;\; P(x) \land P(B)$ \\
                                           &                              &                                                          & $\forall x \;\; P(x) \land P(\tred{x, }\;B)$        \\\cdashline{3-4}
                                           & Insert Negation              & $P(A) \land P(B) \land P(C)$                             & $P(A) \land \tred{\neg(}P(B) \land P(C)\tred{)}$  \\\cdashline{3-4}
                                           & Insert  Formula              & $P(A) \land P(B)$                                        & $P(A) \land P(B) \;\tred{\to R(C)}$                 \\ \midrule
        \multirow{4}{*}{Delete}            & \multirow{2}{*}{Delete Term} & $\tblue{\forall x} \; \forall y \;\; P(x) \land R(x, y)$ & $\forall y \;\; P(x) \land R(x, y)$               \\
                                           &                              & $\forall x \; \forall y \;\; P(x) \land R(\tblue{x, }\;y)$ & $\forall x \; \forall y \;\; P(x) \land R(y)$     \\\cdashline{3-4}
                                           & Delete Negation              & $\tblue{\neg(}P(A) \land P(B)\tblue{)}$                  & $\neg(P(A) \land P(B))$                           \\\cdashline{3-4}
                                           & Delete Formula               & $P(A) \;\tblue{\land}\; \tblue{P(B)}\; \land P(C)$                     & $P(A) \land P(C)$                                 \\ \bottomrule
        \end{tabular}%
        }
    \end{minipage}    
    \begin{minipage}{0.4\textwidth}
        \centering
        \includegraphics[trim={0 6.7cm 14cm 0},clip,width=\textwidth]{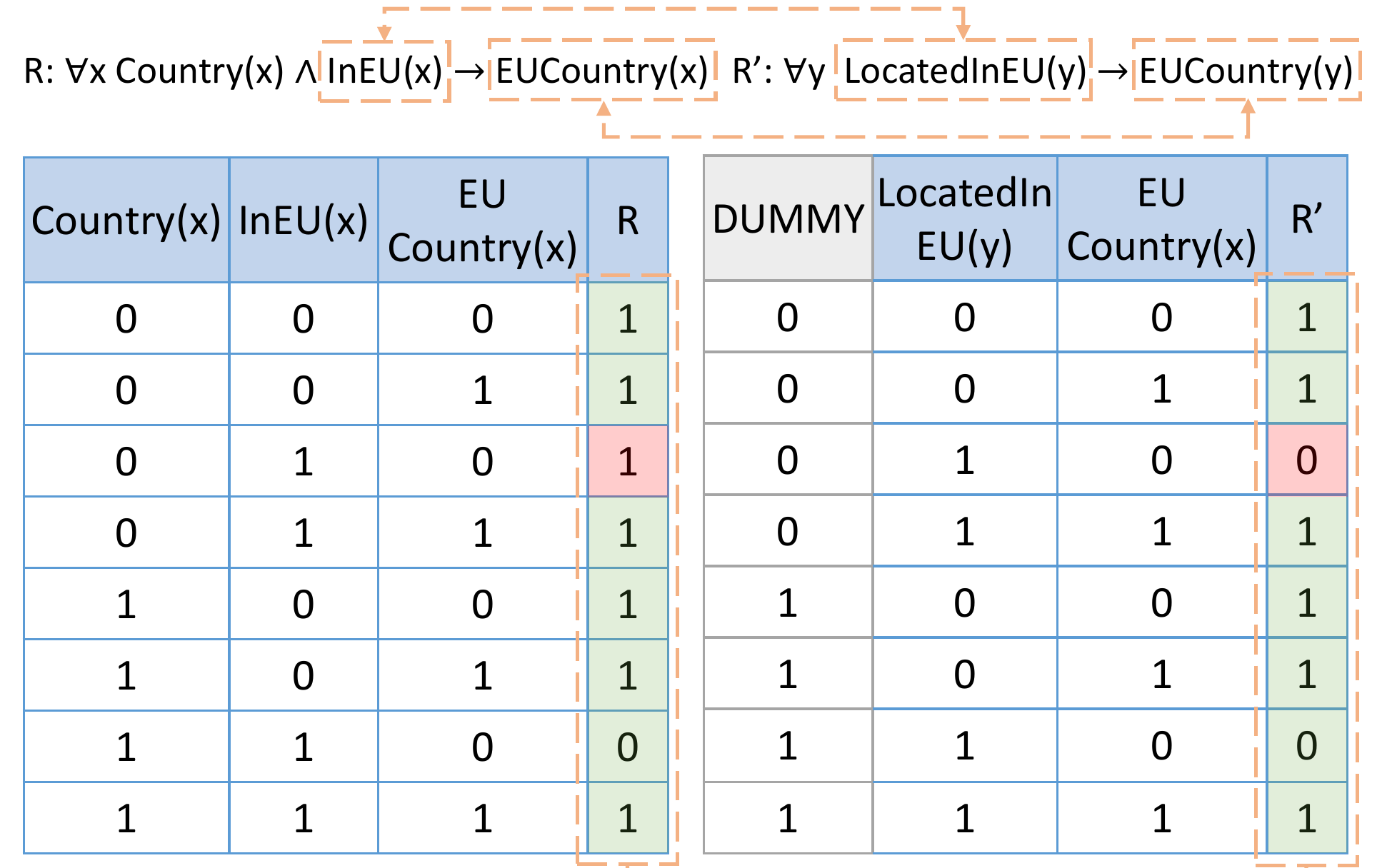}
        \vspace{-.6cm}
        \captionof{figure}{Example of computing the Logical Equivalence score, 7/8=0.875.}
        \label{fig:truth-table-le}
    \end{minipage}    
\end{minipage}
\end{figure*}

Since we do not have the ground-truth CoT steps $\hat{\cX}_\Delta$ for the real output $\hat{\vx}_\text{FOL}$, we generate synthetic steps and the output sequence by randomly perturbing the FOL rules in \dataset.

We consider three types of atomic perturbations: \textit{label change}, \textit{insert}, and \textit{delete}. As shown in Table~\ref{tab:perturb-operations}, label change can be conducted on any terms or logic operators in a FOL rule; insert operation is applicable to term, negation, and formula; and delete operation can be considered as the inverse of insertion.
Note that, we restrict the perturbations to only produce valid rules. The reasons are two-fold: (1) the invalid rule space is effectively the space of all possible strings which is prohibitive to explore; and (2) we found GPT-3.5 rarely generates syntactically invalid rule, thus, limiting the synthetic data in the valid rule space will already cover a wide range of the actual GPT-3.5 outputs.

\textbf{Perturbation process}.
Given a ground-truth pair $\anB{\vx_\text{NL}, \vx_\text{FOL}}$, a parser (Appendix~\ref{app:dataset_app}) will parse $\vx_\text{FOL}$ into an abstract syntax tree (AST). We randomly perturb the AST with atomic operations in Table~\ref{tab:perturb-operations} and for $N_\text{Perturb}$ times. Here, $N_\text{Perturb}$ is also picked randomly from a list of numbers, and in the experiments, we set it to $\{0,1,2,...,10\}$. In the case $N_\text{Perturb}=0$, the perturbed rule remains the same as the ground truth and the CoT step is simply ``No changes needed''; this is effectively a negative example that penalizes the model for over-correcting. During training, we found \method still tends to over-correct the samples as negative samples by default account for around 10\% of the data, so we manually set the probability of negative sample generation to 0.2.

\textbf{Iterative correction}.
Depending on the capacity of the LLM, it might be difficult for the model to learn to output many steps (say 10) within one generation.
We propose to break down the correction into multiple generations, where the model is tasked to output at most $N_\text{Correct}$ steps of correction given the perturbed rule and the previous corrections up to $N_\text{Perturb} - N_\text{Correct}$ steps. For example, the right column of Figure~\ref{fig:three_modes} shows an iterative correction sample: it requires total $N_\text{Perturb}=4$ steps to correct the rule, but we picked a correction steps of $N_\text{Correct}=2$; this means the perturbed rule together with the previous two steps are treated as input and the last two steps are the output.
Similar to $N_\text{Perturb}$, we randomly choose $N_\text{Correct}$ from a list, where we set it to $\{0,1,2,3\}$. And apparently, $N_\text{Correct}$ should be no greater than the total steps $N_\text{Correct} = \min(N_\text{Correct}, N_\text{Perturb})$.

\textbf{SFT for CoT correction}.
For this \textbf{(T3)} task, we generate the synthetic dataset consisting of 150K examples in the form of $\anB{\vx_\text{NL}, \vx_\text{FOL}, \hat{\cX}_{\Delta,\text{prev}}, \hat{\cX}_{\Delta,\text{corr}}, \hat{\vx}_\text{FOL}}$ using the above method, where $\hat{\cX}_{\Delta,\text{prev}}, \hat{\cX}_{\Delta,\text{corr}}, \hat{\vx}_\text{FOL}$ are the previous correction steps, target correction steps and the perturbed FOL rule respectively. We then fine-tune the LLaMA-7B model with LoRA again using the standard autoregression objective: 
the input is $[\vx_\text{NL}, \hat{\vx}_\text{FOL}, \hat{\cX}_{\Delta,\text{prev}}]$ 
and the output is $[\hat{\cX}_{\Delta,\text{corr}}, \vx_\text{FOL}]$.

\subsubsection{RLHF for CoT correction}

With $\textbf{(T3)}$ SFT CoT correction, we enable the model to generate intermediate correction steps for synthetic data. Now, we train the model to correct the actual outputs from GPT-3.5, which is \textbf{(T4)} RLHF CoT Correction task.

\textbf{Why do we need RLHF?}
Note that to achieve this goal for \textbf{(T4)}, we can no longer use the autoregression objective as in \textbf{(T1)}, \textbf{(T2)}, or \textbf{(T3)}, since we still do not have the ground-truth CoT steps for GPT-3.5 outputs.
However, on the other hand, we can still compare the final corrected rule to the ground-truth rule and measure how close they are.
And this gives rise to an RL approach to the problem.
Formally, let $\ttt{RM}: \cX \times \cX \mapsto [0, 1]$ be a function that maps a pair of FOL sequences, $\vx_\text{FOL}$ and $\vx_\text{FOL}'$, to a scalar score representing the pair similarity, our objective can be formalized as maximizing the score (effectively the expected return in RL),
\begin{align}\label{eq:rlhf-obj}
    \max_\pi \ttt{RM}(\vx_\text{FOL}, \vx_\text{FOL}'), \;\text{where}\; \vx_\text{FOL}' \sim \pi_\theta(\vx_\text{FOL}, \hat{\cX}_{\Delta,\text{corr}}| \vx_\text{NL}, \hat{\cX}_{\Delta,\text{prev}}, \hat{\vx}_\text{FOL}),
\end{align}
 for all tuples $\anB{\vx_\text{NL}, \vx_\text{FOL}, \hat{\vx}_\text{FOL}}$ in \dataset via a policy $\pi_\theta(\vx_\text{FOL}, \hat{\cX}_{\Delta,\text{corr}}| \vx_\text{NL}, \hat{\cX}_{\Delta,\text{prev}}, \hat{\vx}_\text{FOL})$ which is exactly the autoregressive model we trained in \textbf{(T3)} and would like to fine-tune in \textbf{(T4)}.
With objective~\eq{eq:rlhf-obj}, task \textbf{(T4)} is now similar to the RLHF proposed in InstructGPT~\citep{ouyang2022training} with the only difference being the reward model \ttt{RM}, where in our case, \ttt{RM} is a logical equivalence solver (\S\ref{sec:fol-eval-reward-le}) instead of a language model.

\textbf{Training process}.
In \textbf{(T4)} RLHF CoT correction, we fine-tune the \method model obtained in \textbf{(T3)} SFT CoT correction via RLHF. For every tuple $\anB{\vx_\text{NL}, \vx_\text{FOL}, \hat{\vx}_\text{FOL}}$, we let the model to continuously generate the corrections
$[
\anB{\vx_\text{FOL}'^{(1)}, \hat{\cX}_{\Delta,\text{corr}}^{(1)}},
\anB{\vx_\text{FOL}'^{(2)}, \hat{\cX}_{\Delta,\text{corr}}^{(2)}},
...
]$
until the model outputs ``No changes needed'' in the CoT steps or hits the token limit; the previous correction $\hat{\cX}_{\Delta,\text{prev}}$ is set to empty initially and we update it with the output steps in every generation. In other words, at iteration $(t)$, the previous correction is $\hat{\cX}_{\Delta,\text{prev}} = [
    \hat{\cX}_{\Delta,\text{corr}}^{(1)},
    \hat{\cX}_{\Delta,\text{corr}}^{(2)},
    ...,
    \hat{\cX}_{\Delta,\text{corr}}^{(t-1)},
]$.
For every generated text FOL at iteration $(t)$, we collect the experience tuple $\anB{\vx_\text{FOL}'^{(t)},\vx_\text{NL}, \hat{\cX}_{\Delta,\text{prev}}, \hat{\vx}_\text{FOL}, r^{(t)}}$ where $r^{(t)}=\ttt{RM}(\vx_\text{FOL}, \vx_\text{FOL}'^{(t)})$, and once enough experience is collected, we update the model parameter $\theta$ via PPO~\citep{schulman2017proximal}.





\subsubsection{FOL evaluation and reward design}
\label{sec:fol-eval-reward-le}

The last component for \textbf{(T4)} is the reward model \ttt{RM}. This requires a metric that measures the similarity between two text FOLs $\vx_\text{FOL}$, $\vx_\text{FOL}'$ to be implemented, and the metric should take into account the scenarios mentioned in challenge \textbf{(C2)}.

\textbf{Logical equivalence (LE)}. 
We propose to measure the logical equivalence between the rules by matching their truth tables and computing the overlap ratio.
We introduce this with a running example in Figure~\ref{fig:truth-table-le}.
Specifically, let $R$ and $R'$ be the two rules parsed from the text $\vx_\text{FOL}$ and $\vx_\text{FOL}'$. We identify the set of literals in each rule $\cP = [p_1, p_2,...]$ and $\cQ = [q_1, q_2,...]$. 
In the case of Figure~\ref{fig:truth-table-le}, $\cP = [\ttt{Country}(x), \ttt{InEU}(x), \ttt{EUCountry}(x)]$ and $\cQ = [\ttt{LocatedInEU(y)}, \ttt{EUCountry}(y)]$.
One can consider the set of literals as an array of Boolean variables, and the FOL as a circuit that takes in the Boolean values and outputs a single Boolean value. Therefore, we can represent a FOL with a truth table that enumerates all possible inputs and the resulting outputs.
And to compare $R$ and $R'$, we count the number of configurations that match and divide it by the total number of configurations; this yields a score in $[0,1]$.
In Figure~\ref{fig:truth-table-le}, this is $7/8 = 0.875$.
The main issue with this approach is finding the right input bindings between $\cP$ and $\cQ$, and dealing with the case where the numbers of inputs are different (i.e., $|\cP|\neq|\cQ|$).
We solve this by finding the binding that gives the highest LE score via greedy search and filling the rest of the missing inputs with dummy inputs.
In Figure~\ref{fig:truth-table-le}, $\ttt{InEU}(x)$ binds to $\ttt{LocatedInEU(y)}$ and $\ttt{EUCountry}(x)$ binds to $\ttt{EUCountry}(y)$; and we fill in a dummy in $\cQ$ to match $\ttt{Country}(x)$ in $\cP$.
We leave more details in Appendix~\ref{app:le_computation}.

\textbf{Reward design}.
We use the LE score as the main source of the reward. However, we also want the model to extract the right predicate and entity names from the NL statement. We incorporate this aspect by computing the BLEU score between the text $\vx_\text{FOL}$ and $\vx_\text{FOL}'$ with a specialized FOL tokenizer. We set the final reward as the mixture of the two: 
$
\ttt{RM}(\vx_\text{FOL}$, $\vx_\text{FOL}') = 
\omega * \text{LE}(R, R') + (1 - \omega) * \text{BLEU}(\vx_\text{FOL}$, $\vx_\text{FOL}')
$
, where $\omega$ is the mixing ratio and in experiments we set it to 0.7.

\section{Experiments}
\label{sec:experiments}

We address the following questions in the experiment section:
(\textbf{Q1}) How good is \dataset? Can we train a strong NL-FOL translation model with a ``silver-labels-only'' dataset?
(\textbf{Q2}) How well does the \method perform in direct translation mode and CoT correction mode?
(\textbf{Q3}) How do the CoT corrections influence the performance of \method?




\textbf{Dataset}.
We use the entire \dataset as the training set for \textbf{(T1)-(T4)}; we also include 1K pairs from the training set of LogicNLI since it has a different rule distribution where rules are mostly grounded rules (i.e., many of them do not contain any variables) instead of FOL rules. We evaluate the LLMs on the full FOLIO dataset and the test set of LogicNLI.

\textbf{Training, generation, and hardware settings}.
For all training tasks, we fine-tune \method using LoRA with rank=16, $\alpha=16$, and dropout 0.05 on all the LLaMA-7B attention weights.
We use the AdamW optimizer~\citep{loshchilov2017decoupled} with $lr=0.0003$.
For the generation, we use a cutoff length of 256 for \textbf{(T1)} and \textbf{(T2)}; and 1024 for \textbf{(T3)} and \textbf{(T4)}, where 748 and 256 are allocated for the input prompt and output sequences respectively.
All experiments are conducted on a Xeon 6140 machine with 256G RAM and a single V100 GPU (Detailed settings at Appendix~\ref{app:exp-setting}).


\textbf{Metrics}.
We evaluate the translated and the final corrected FOL rules with two metrics: FOL BLEU score and FOL logical equivalence (LE) score (\S\ref{sec:fol-eval-reward-le}).


\begin{figure*}[t]
    \begin{minipage}{\textwidth}
    \begin{minipage}{0.54\textwidth}
        \vspace{0pt}
        \centering
        \captionof{table}{BLEU and the logical equivalence (LE) scores of \method and GPT models on LogicNLI and FOLIO. Direct translation using \method outperforms GPT-3.5 and CoT correction achieves a similar performance as 5-shot GPT-4.}
        \label{tab:main-res-table}
        \resizebox{\columnwidth}{!}{%
        \begin{tabular}{@{}lcclcc@{}}
        \toprule
        \multicolumn{1}{c}{\multirow{2}{*}{Methods}} & \multicolumn{2}{c}{LogicNLI} &  & \multicolumn{2}{c}{FOLIO} \\ \cmidrule(lr){2-3} \cmidrule(l){5-6} 
        \multicolumn{1}{c}{}                         & FOL BLEU       & FOL LE      &  & FOL BLEU     & FOL LE     \\ \midrule
        GPT-3.5 0-shot                                & 0.584          & 0.589       &  & 0.248        & 0.429      \\
        GPT-3.5 5-shot                                & 0.905          & 0.918       &  & 0.341        & 0.767      \\
        GPT-4 0-shot                                  & 0.740          & 0.863       &  & 0.372        & 0.799      \\
        GPT-4 5-shot                                  & 0.913          & 0.989       &  & 0.400        & 0.855      \\ \midrule
        Direct Translation                           & 0.926          & 0.965       &  & 0.372        & 0.818      \\
        Naive Correction                             & 0.934          & 0.970       &  & 0.373        & 0.840      \\
        SFT CoT Correction                           & 0.663          & 0.830       &  & 0.332        & 0.730      \\
        RLHF CoT Correction                          & 0.935          & 0.978       &  & 0.385        & 0.849      \\ \bottomrule
        \end{tabular}%
        }
    \end{minipage}    
    \begin{minipage}{0.45\textwidth}
        \centering
        \captionof{table}{RLHF CoT correction performance vs. Max \# generations.}
        \begin{minipage}[t]{0.9\textwidth}
            \centering            
            \label{tab:steps-choice}
            \resizebox{\columnwidth}{!}{%
            \begin{tabular}{@{}llclclclcl@{}}
            \toprule
             & \multicolumn{1}{c}{\multirow{2}{*}{Metrics}} & \multicolumn{8}{c}{Max \# Generations} \\ \cmidrule(l){3-10} 
             & \multicolumn{1}{c}{}                         & 1   &    & 3   &    & 5   &    & 10     &   \\ \midrule
             & FOL BLEU                                     & 0.361   &    & 0.370   &    & 0.384   &    & 0.385  &   \\
             & FOL LE                                       & 0.815   &    & 0.841   &    & 0.846   &    & 0.849  &   \\ \bottomrule
            \end{tabular}%
            }
        \vspace{.2cm}
        \end{minipage}        
        \begin{minipage}[b]{0.49\textwidth}
            \centering
            \includegraphics[trim={0 0cm 1.5cm 1.3cm},clip,width=\textwidth]{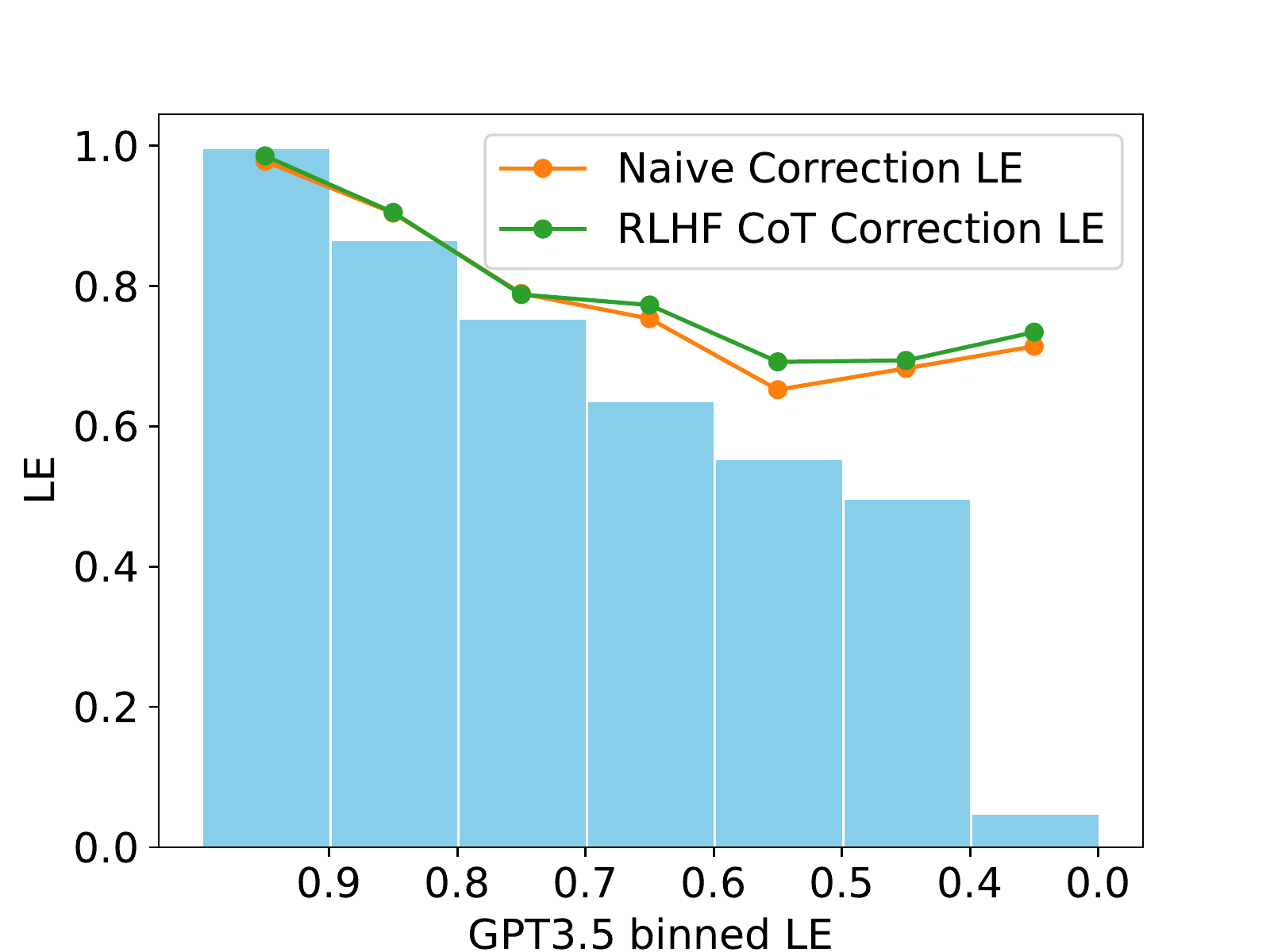}
        \end{minipage}
        \hfill
        \begin{minipage}[b]{0.49\textwidth}
            \centering
            \includegraphics[trim={0 0cm 1.5cm 1.3cm},clip,width=\textwidth]{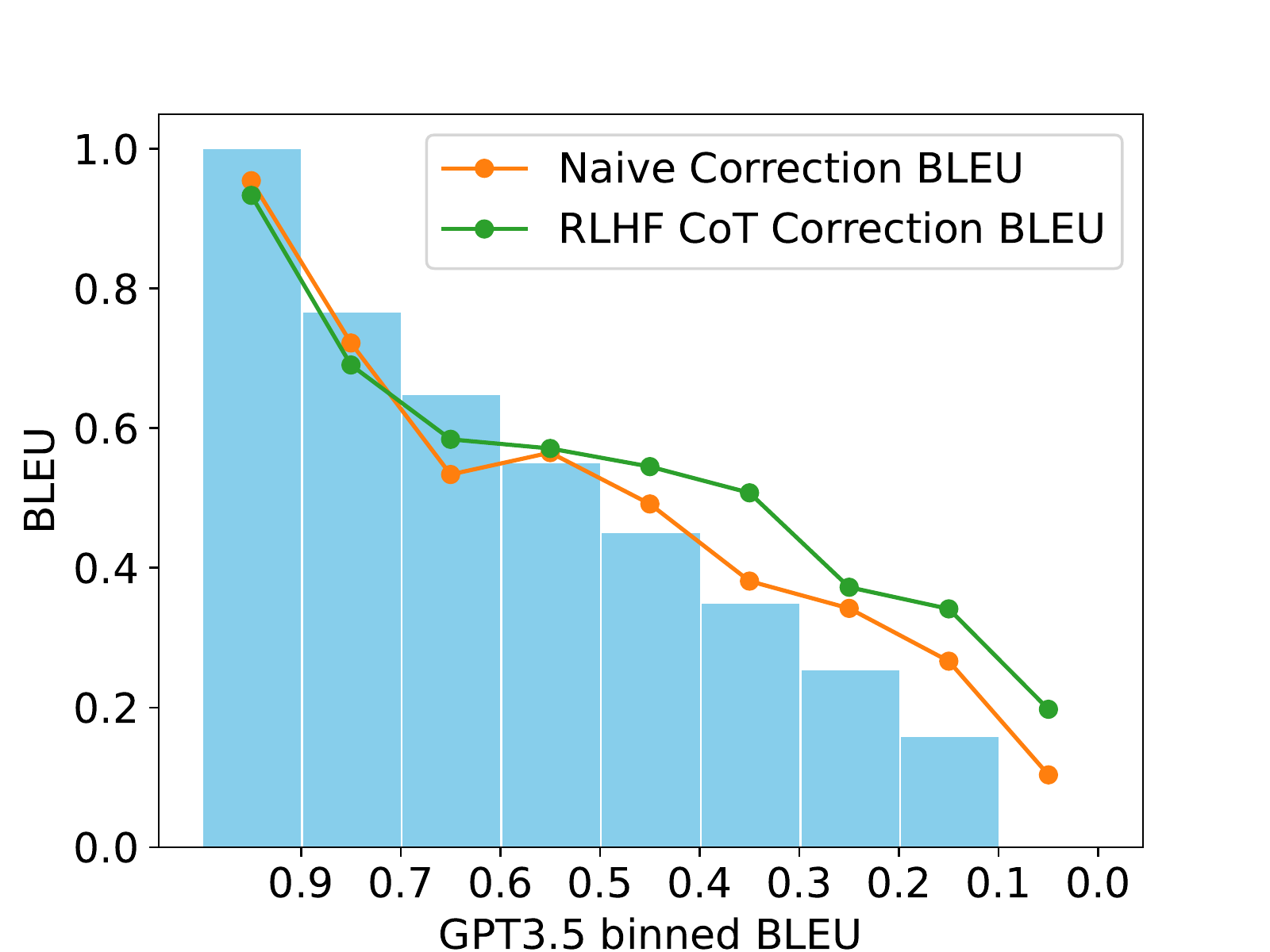}
        \end{minipage}
        \captionof{figure}{\method correction performance averaged over corresponding GPT-3.5 LE and BLEU scores.}
        \label{fig:binned-scores}
    \end{minipage}    
\end{minipage}
\end{figure*}
\begin{figure}[t]
    \centering
    \includegraphics[trim={0 7.8cm 2cm 0},clip,width=0.95\linewidth]{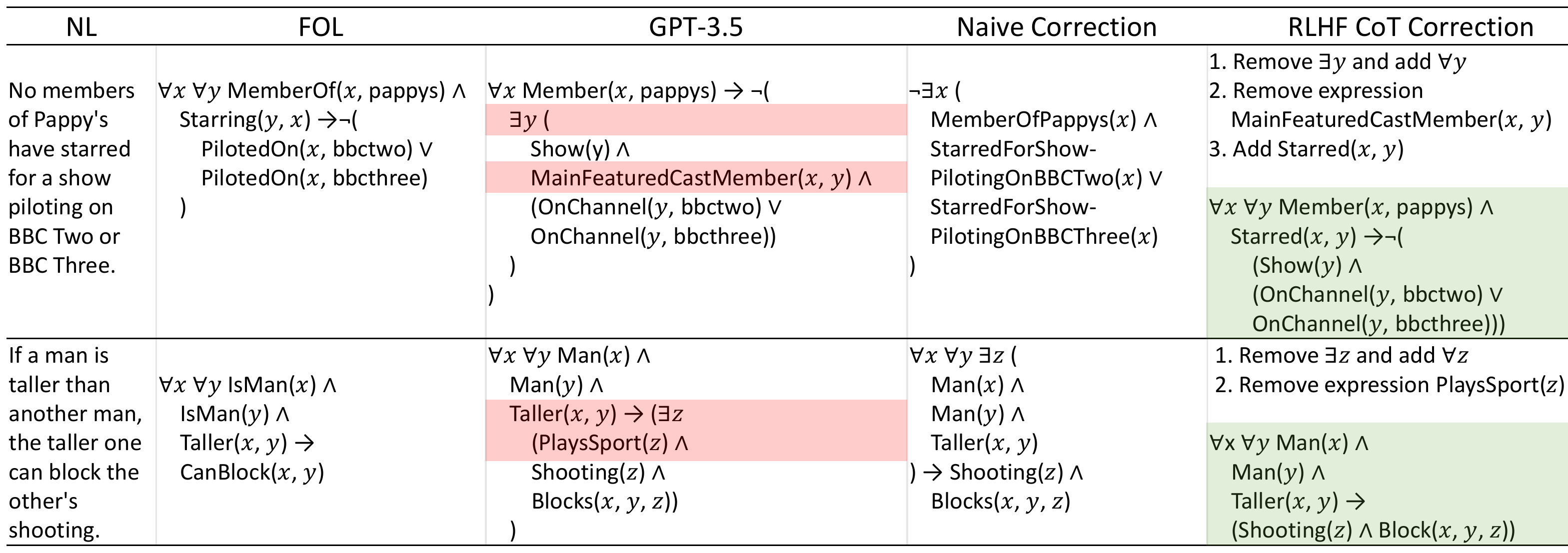}
    \caption{Examples of correcting GPT-3.5's output via naive and RLHF CoT correction.}
    \label{fig:correction-examples}
\end{figure}

\subsection{Results}

Table~\ref{tab:main-res-table} shows the results of \method and GPT models on the LogicNLI and the FOLIO dataset.
In general, we found that 5-shot GPT-4, as the most powerful LLM to date, achieves the best performance for both benchmarks.
On the other hand, \method outperforms GPT-3.5 models in both translation and correction modes, and the best performance is achieved by RLHF CoT correction which leads to a GPT-4 level performance. This suggests that \dataset---while being a silver label dataset---can indeed produce an LLM comparable to GPT-4 on a gold set, which addresses the question \textbf{(Q1)}.
Benchmark-wise, all methods achieve near-perfect results 
on LogicNLI except for 0-shot GPT-3.5, which has trouble generating syntactically valid rules due to the lack of examples. This is because LogicNLI is synthetically generated and the rules all share a similar FOL template.
On the other hand, FOLIO is more challenging as they are expert-written.




\subsection{Analysis}

\textbf{Translation vs. Correction}.
Table~\ref{tab:main-res-table} suggested that the correction mode \method leads to better performance than the direct translation mode. This confirms our intuition in \S\ref{sec:naive-method} and addresses the question \textbf{(Q2)}.
More importantly, these results suggest a new paradigm of future LLM development: by training a local LLM on the output of a more powerful model, one can conduct in-depth customization on the model behavior while still leveraging the generalizability of the powerful LLMs for heavy lifting.
This paradigm is beneficial as GPT-3.5 and GPT-4 nowadays do not support fine-tuning and have a limited context window for customization.

\textbf{Effect of CoT correction}.
To see how and why CoT correction improves performance, we compare the 
\textbf{(T2)} naive and the \textbf{(T4)} CoT correction performance on samples grouped by their ``difficulty'' level. To do this, we group samples by the GPT-3.5's LE and BLEU scores into several bins (e.g., [1.0-0.9], [0.9-0.8] and etc.). Within each bin, we average the scores of GPT-3.5, \textbf{(T2)}, and \textbf{(T4)}. The results are shown in Figure~\ref{fig:binned-scores}.
And correction examples are shown in Figure~\ref{fig:correction-examples}.
We find the correction leads to a better performance generally by improving the difficult examples where GPT-3.5 fails significantly.
The same trend is also present between \textbf{(T2)} and \textbf{(T4)}, where CoT leads to better performance, especially on the BLEU score. We conjecture this is because CoT elicits the intermediate steps making it easy to find the right predicate and entity names.

\textbf{Effect of CoT steps}.
We study the effect of the CoT steps by varying the maximum number of allowed generations on a single sample, which effectively limits the number of CoT steps that could be made by the model. The results are shown in Table~\ref{tab:steps-choice}.
We found the performance saturated quickly starting from a max of three generations. For the one-generation case, it is slightly worse than the naive correction counterpart due to only a limited number of corrections could be made.


\section{Conclusion}
We present \method, the first specialized LM for the NL-FOL translation task. We release a high quality dataset of 34K sentence-level NL-FOL pairs collected from GPT-4, used for fine-tuning \method. \method with only 7B parameters shows competitive performance with GPT-4, while outperforming GPT-3.5 on challenging held-out NL-FOL benchmark. Through a novel SFT+RLHF training framework, we equip \method with step-by-step corrective capability, allowing it to  consistently correct its own outputs, as well as outputs from a large LM (i.e., GPT-3.5).  





\clearpage
\newpage
\bibliographystyle{plainnat}
\bibliography{ref}



\clearpage
\newpage


\appendix

\section{Appendix}




In the following, we provide further details about the dataset creation, logical equivalence computation and experimental settings.

\section{\dataset Dataset Creation Details}
\label{app:dataset_app}

\begin{figure}[t]
    \centering
    \includegraphics[trim={0 0 9cm 0},clip,width=\linewidth]{figs/sankey_diagram.pdf}
    \caption{Top 200 frequent FOL term pairs in \dataset. Many terms are associated with a wide range of other terms, which suggests the rules are semantically and contextually diverse.}
    \label{fig:sankey}
\end{figure}
\begin{figure*}[h]
    \begin{minipage}{\textwidth}
    \begin{minipage}[b]{0.49\textwidth}
        \centering
        \includegraphics[trim={1.3cm 0cm 0.7cm 1.3cm},clip,width=\textwidth]{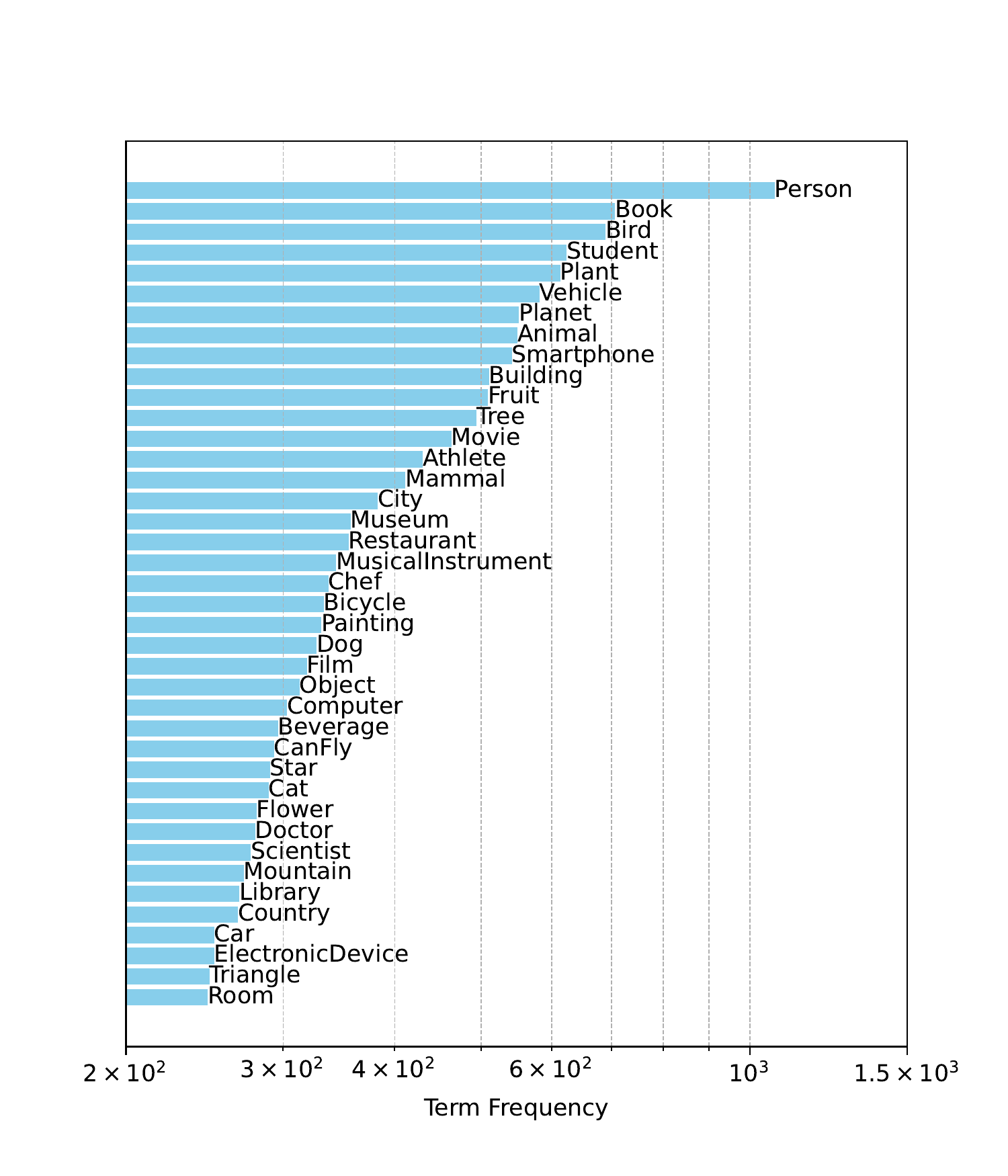}
        \captionof{figure}{Top 40 frequent FOL terms (\dataset).}
        \label{fig:term-freq}
    \end{minipage}
    \hfill
    \begin{minipage}[b]{0.49\textwidth}
        \centering
        \includegraphics[trim={0 0cm 1.5cm 1.3cm},clip, width=\textwidth]{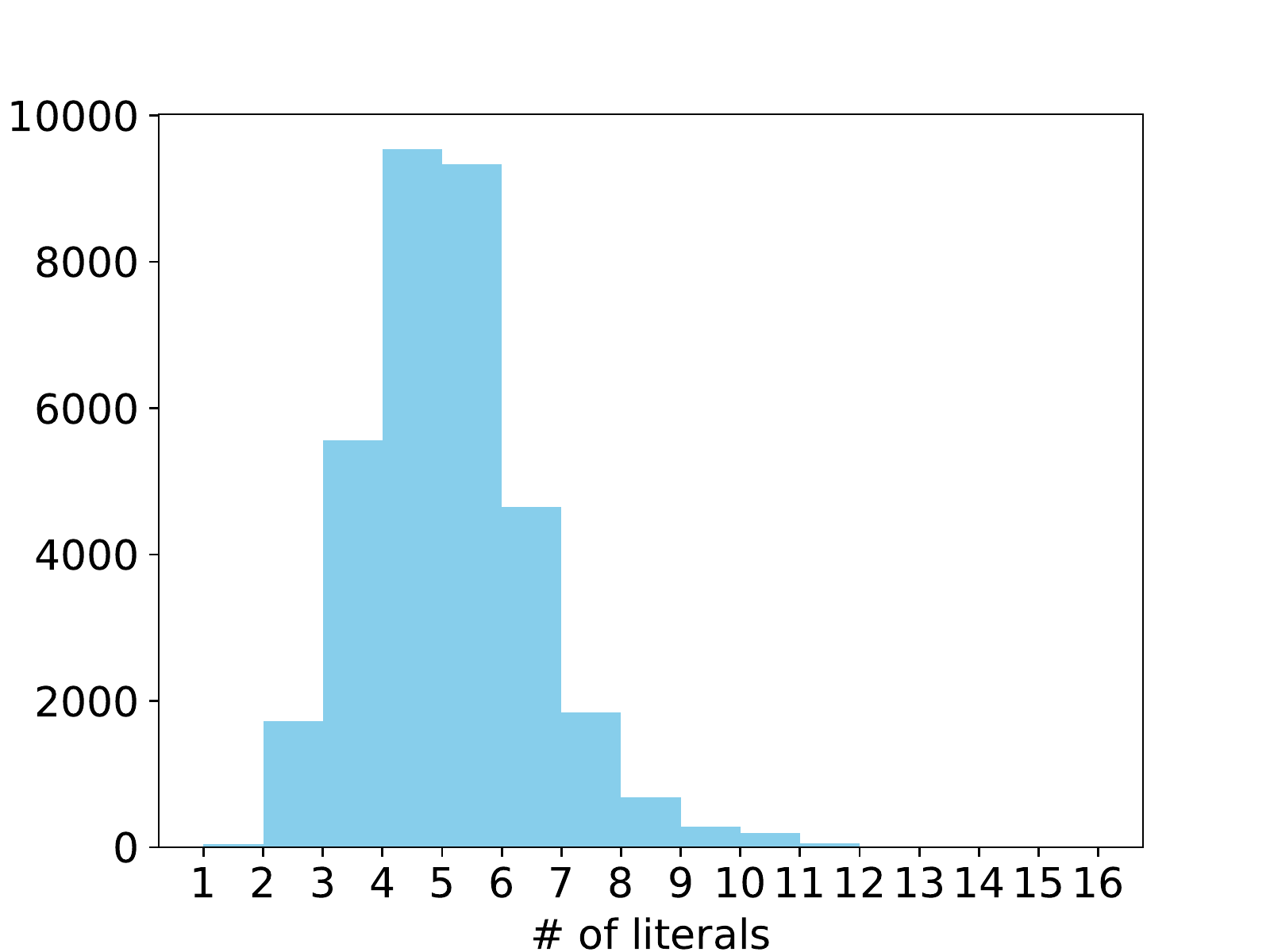}
        \captionof{figure}{Literal frequency distribution (\dataset).}
        \label{fig:literal-freq}
    \end{minipage}
\end{minipage}
\end{figure*}
\begin{table}[h]
\centering
\caption{List of prompt templates used for prompting GPT4 for NL-FOL pairs.}
\label{tab:prompt-table}
\resizebox{\columnwidth}{!}{%
\begin{tabular}{@{}ll@{}}
\toprule
\begin{tabular}[c]{@{}l@{}}System\\ prompt\end{tabular}               & \begin{tabular}[c]{@{}l@{}}I want to create a dataset for translating natural language (NL) statements into first-order logic (FOL) rules. \\ You will help me to create a diverse set of NL-FOL pairs.\\ \\ For natural language (NL) generation, you should:\\     1. Come up with a statement stating either complex or simple real-world commonsense facts\\     2. The statements are meaningful, and diverse from each other\\ \\ For FOL rule generation:\\     1. You SHOULD USE the following logical operators: $\oplus$ (either or), $\lor$ (disjunction), \\         $\land$ (conjunction), $\to$ (implication), $\forall$ (universal), $\exists$ (existential), $\neg$ (negation), \\         $\leftrightarrow$ (equivalence)\\     2. You *SHOULD NEVER USE* the following symbols for FOL: "\!", "$\neq$", "\%", "="\\     3. The literals in FOL SHOULD ALWAYS have predicate and entities, e.g., "Rounded(x, y)" or "City(guilin)";\\         expressions such as "y = a $\lor$ y = b" or "a $\land$ b $\land$ c" are NOT ALLOWED\\     4. The FOL rule SHOULD ACCURATELY reflect the meaning of the NL statement\\     5. You SHOULD ALWAYS put quantifiers and variables at the beginning of the FOL\\     6. You SHOULD generate FOL rules with either: (1) no variables; (2) one variable "x"; (3) two variables "x", \\         "y"; or (4) three variables "x", "y" and "z"\\ \\ Generation Format: you SHOULD ALWAYS generate the NL and FOL pairs in the following format\\ """\\ --- NL:\\ \{your generated NL\}\\ ---\\ --- FOL:\\ \{your generated FOL\}\\ ---\\ """\end{tabular} \\ \midrule
\begin{tabular}[c]{@{}l@{}}Few-shot \\ examples\\ prompt\end{tabular} & \begin{tabular}[c]{@{}l@{}}--- NL:\\ If someone is entire, then he is not serious, and vice versa. \\ --- FOL:\\ $\exists$x entire(x) $\leftrightarrow$ $\neg$serious(x)\\ \\ --- NL: \\ If there is at least one people who is both not excited and not timid, then Jonathan is elderly.\\ --- FOL:\\ $\forall$x ($\neg$excited(x) $\land$ $\neg$timid(x)) $\to$ elderly(Jonathan)\\ \\ --- NL: \\ Someone who is eithor not fresh or entire is always not serious.\\ --- FOL:\\ $\forall$x ($\neg$concerned(x) $\lor$ fresh(x)) $\to$ entire(John)\\ \\ --- NL: \\ If Nathalie is not blue, then Collier is entire.\\ --- FOL:\\ $\neg$blue(Nathalie) $\to$ entire(Collier)\\ \\ --- NL: \\ Someone is courteous and not elderly if and only if he is not excited and not various.\\ --- FOL:\\ $\exists$x (courteous(x) $\land$ $\neg$elderly(x)) $\leftrightarrow$ ($\neg$excited(x) $\land$ $\neg$various(x))\end{tabular}                                                                                                                                                                                                                                                                                                                                                                                                                                                                                                                                                                                                                                                                               \\ \midrule
\begin{tabular}[c]{@{}l@{}}Negative \\ N-gram\\ prompt\end{tabular}   & \begin{tabular}[c]{@{}l@{}}They DO NOT involve concepts and terms (and the synonyms) such as "considered","person","either","water",\\ "if it has","if it is","it has a","is considered a","A person is"\end{tabular}                                                                                                                                                                                                                                                                                                                                                                                                                                                                                                                                                                                                                                                                                                                                                                                                                                                                                                                                                                                                                                                                                                                                                                                                                                                                                                                                                                                         \\ \midrule
\multirow{2}{*}{FOL prompts}                                          & They are {[complex | simple]} statements involving at least {[1 | 2 | 3]} logical variables                                                                                                                                                                                                                                                                                                                                                                                                                                                                                                                                                                                                                                                                                                                                                                                                                                                                                                                                                                                                                                                                                                                                                                                                                                                                                                                                                                                                                                                                                                                   \\ \cmidrule(l){2-2} 
                                                                      & The statement involves diverse logical operators such as logical negation, logical xor and disjunction                                                                                                                                                                                                                                                                                                                                                                                                                                                                                                                                                                                                                                                                                                                                                                                                                                                                                                                                                                                                                                                                                                                                                                                                                                                                                                                                                                                                                                                                                                        \\ \midrule
\begin{tabular}[c]{@{}l@{}}Break-down \\ prompt\end{tabular}          & {[IMPORTANT]} AVOID making long predicate names like "MoonShinesAtNight","SunShinesDuringDay"                                                                                                                                                                                                                                                                                                                                                                                                                                                                                                                                                                                                                                                                                                                                                                                                                                                                                                                                                                                                                                                                                                                                                                                                                                                                                                                                                                                                                                                                                                                 \\ \bottomrule
\end{tabular}%
}
\end{table}

\subsection{Data collection}

\textbf{Prompt table}.
Table~\ref{tab:prompt-table} shows the prompt templates used for prompt generation.


\subsection{FOL parsing and verification}

\begin{figure*}[h]
    \begin{minipage}{\textwidth}
    \begin{minipage}[t]{\textwidth}
        \centering
        \includegraphics[width=\textwidth]{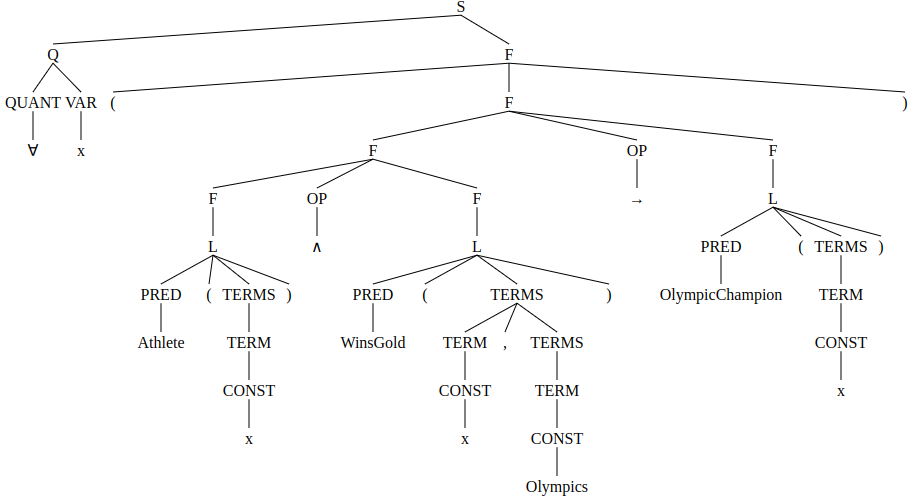}
        \captionof{figure}{CFG parse tree of FOL rule $\forall x (\ttt{Athlete}(x) \land \ttt{WinsGold}(x, \ttt{Olympics}) \to \ttt{OlympicChampion}(x))$.}
        \label{fig:example_parse_tree_1}
    \end{minipage}
    \begin{minipage}[t]{\textwidth}
        \centering
        \includegraphics[width=0.65\textwidth]{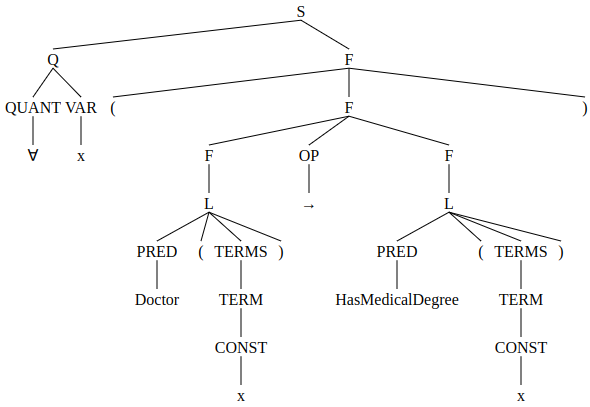}
        \captionof{figure}{CFG parse tree of FOL rule $\forall x (\ttt{Doctor}(x) \to \ttt{HasMedicalDegree}(x))$.}
        \label{fig:example_parse_tree_2}
    \end{minipage}
\end{minipage}
\end{figure*}

\textbf{FOL CFG grammar}.
We define the FOL with the following CFG grammar:

\ttt{S} -> \ttt{F}   |   \ttt{Q}   \ttt{F}

\ttt{Q} -> \ttt{QUANT}   \ttt{VAR} | \ttt{QUANT} \ttt{VAR} \ttt{Q}

\ttt{F} -> `$\neg$' `(' \ttt{F} `)' | `(' \ttt{F} `)' | \ttt{F} \ttt{OP} \ttt{F} | \ttt{L}

\ttt{OP} -> `$\oplus$' | `$\lor$' | `$\land$' | `$\to$' | `$\leftrightarrow$'

\ttt{L} -> `$\neg$' \ttt{PRED} `(' \ttt{TERMS} `)' | \ttt{PRED} `(' \ttt{TERMS} `)'

\ttt{TERMS} -> \ttt{TERM} | \ttt{TERM} `,' \ttt{TERMS}

\ttt{TERM} -> \ttt{CONST} | \ttt{VAR}

\ttt{QUANT} -> `$\forall$' | `$\exists$'

Note that, for \ttt{PRED}, \ttt{CONST}, and \ttt{VAR} they have corresponding production rules generated for each FOL rule example. For example, for rule
``$\forall x ((\ttt{Person}(x) \land \ttt{Drinks}(x)) \to \ttt{DependentOn}(x, \ttt{Caffeine}))$'',
the production rules are
\begin{align*}
    &\ttt{PRED} \to \ttt{`Person'} \;|\; \ttt{`Drinks'} \;|\; \ttt{`DependentOn'} \\
    &\ttt{CONST} \to \ttt{`Caffeine'} \\
    &\ttt{VAR} \to \text{`}x\text{'}.
\end{align*}

We show two example parse trees in Figure~\ref{fig:example_parse_tree_1} and Figure~\ref{fig:example_parse_tree_2}.


\subsection{\dataset statistics}

\textbf{General statistics}.
Figure~\ref{fig:term-freq} and Figure~\ref{fig:literal-freq} show the top 40 frequent FOL terms and the literal count distribution in \dataset.

\textbf{Frequent FOL term pairs}.
Figure~\ref{fig:sankey} shows the top 200 frequent FOL term pairs in \dataset.

\section{Computing Logical Equivalence and BLEU Score}
\label{app:le_computation}

\textbf{logical equivalence}.
To train and evaluate \method, we compute the logical equivalence score (LE) that measures the similarity between two rules $R$ and $R'$. The computation is done in three steps: (1) finding the literals of $R$ and $R'$, that is $\cP=[p_1,p_2,...]$ and $\cQ=[q_1,q_2,...]$; (2) binding the literals in $\cP$ to those in $\cQ$ (or vice versa); and (3) generating the truth tables for the binding and computing the score.

Finding the literals of a FOL rule is straightforward after we parse it into a CFG tree: we extract all the subtrees whose root label is \ttt{L} and remove possible duplicate literals.
In the case where the parsing fails, we simply skip the rest of the computation and return a score of zero, as that indicates the rule is syntactically invalid.

The main challenge here is to determine the literal binding between $\cP$ and $\cQ$. Using Figure~\ref{fig:truth-table-le} as the example, $R$ has literals $\cP = [\ttt{Country}(x), \ttt{InEU}(x), \ttt{EUCountry}(x)]$ and $R'$ has literals $\cQ = [\ttt{LocatedInEU(y)}, \ttt{EUCountry}(y)]$. We want to find the one-one matching for each of the literals, such that we can compare the truth tables. First, we address the case where $|\cP| \neq |\cQ|$ by adding \ttt{DUMMY} inputs to the shorter one, and in this example, it is $\cQ$ which becomes $[\ttt{LocatedInEU(y)}, \ttt{EUCountry}(y), \ttt{DUMMY1}]$.
To match the literals, we first determine the matching strategy. Note that there are in total $!|\cQ|$ numbers of bindings (permute $\cQ$ when keeping $\cP$) and there are many strategies to measure the match: for example, one can enumerate all bindings and compute the ``average'' score of all bindings or finding the worst case of the binding.
Here, we choose to find the binding that yields the highest LE score, that is the ``best'' case binding.
To do this, we implement a simplistic greedy search algorithm that iterates over each literal in $\cP$ and finds the closest literal in $\cQ$ in terms of edit distance.
To avoid exponential numbers of bindings, we limit the search depth to 1000.
Finally, given a binding between $\cP$ and $\cQ$, we compute the LE score by comparing the rows in their truth tables as the one shown in Figure~\ref{fig:truth-table-le}.



\textbf{FOL BLEU score}.
We use a specialized tokenizer for computing the FOL BLEU score. This tokenizer splits every quantifier, operator, and term into tokens. The split token sequence is the same as the leave nodes in the CFG parse tree (Figure~\ref{fig:example_parse_tree_1} and Figure~\ref{fig:example_parse_tree_2}) listed in pre-order.

\section{Experimental Settings}
\label{app:exp-setting}

For all training tasks, we fine-tune \method using LoRA with rank=16, $\alpha=16$, and dropout 0.05 on all the LLaMA-7B attention weights ``{[q\_proj,k\_proj,v\_proj,o\_proj]}''.
We use the AdamW optimizer~\citep{loshchilov2017decoupled} with $lr=0.0003$.
For the generation, we use a cutoff length of 256 for \textbf{(T1)} and \textbf{(T2)}; and 1024 for \textbf{(T3)} and \textbf{(T4)}, where 748 and 256 are allocated for the input prompt and output sequences respectively.
For \textbf{(T1-T3)}, the generation uses temperature=0.1, top\_p=0.75, top\_k=40 and num\_beams=1. For \textbf{(T4)}, we adopt the setting suggested in the \href{https://github.com/lvwerra/trl}{TRL} library, which uses top\_k = 0.0, top\_p = 1.0, do\_sample = True and no eos token; this effectively lets the model sample tokens from the logits and always generate to the full length. This generation configuration is needed to compute a valid KL divergence score between the actor model and the reference model (a copy of the same model before training).

Recall that \textbf{(T4)} generates corrections in multiple rounds of generations, where previous corrections are appended to the initial prompt and fed to the model again as the input prompt (Figure~\ref{fig:three_modes}). For all experiments, we set the max rounds of generation to 10, except for Table~\ref{tab:steps-choice} which examines the model's performance by varying the max rounds. Also, we found that GPT-3.5 can sometimes generate syntactically invalid FOL rules which lie outside of rule space simulated in \textbf{(T3)}. We address this by first correcting the GPT-3.5 response with naive correction \textbf{(T2)} and then feeding the output to \textbf{(T4)}.




\end{document}